%% file: main.tex
\documentclass[10pt,twocolumn,letterpaper]{article}

\newcommand{\methodname}{CoA-VLA}
\usepackage[pagenumbers]{iccv} 

\usepackage{amsmath}
\usepackage{amssymb}
\usepackage{mathtools}
\usepackage{amsthm}
\usepackage[dvipsnames]{xcolor}
\definecolor{AliceBlue}{rgb}{0.941, 0.973, 1}  
\usepackage{algorithmic}
\usepackage{epsfig}
\usepackage{graphicx}

\usepackage{booktabs}
\usepackage{multirow}
\usepackage{multicol}


\definecolor{iccvblue}{rgb}{0.21,0.49,0.74}
\usepackage[pagebackref,breaklinks,colorlinks,allcolors=iccvblue]{hyperref}

\def\paperID{11598} 
\def\confName{ICCV}
\def\confYear{2025}

\title{{CoA-VLA: Improving Vision-Language-Action Models via \\ Visual-Textual Chain-of-Affordance}}

\author{
Jinming Li\textsuperscript{1,*,$\star$},
Yichen Zhu\textsuperscript{2,†,$\star$},
Zhibin Tang\textsuperscript{$\star$},
Junjie Wen\textsuperscript{3,$\star$},
Minjie Zhu\textsuperscript{3,$\star$},
Xiaoyu Liu\textsuperscript{1,$\star$},\\
Chengmeng Li\textsuperscript{1,$\star$},
Ran Cheng\textsuperscript{2},
Yaxin Peng\textsuperscript{1,†,$\star$},
Yan Peng\textsuperscript{1},
Feifei Feng\textsuperscript{2}
\\
\small
\textsuperscript{1}Shanghai University, 
\textsuperscript{2}Midea Group, 
\textsuperscript{3}East China Normal University
\\
\footnotesize
* Co-first author, 
† Corresponding Author,
$\star$ Core Contributor
}
\begin{document}
\twocolumn[
\maketitle

{
\centering
\includegraphics[width=0.88\textwidth]{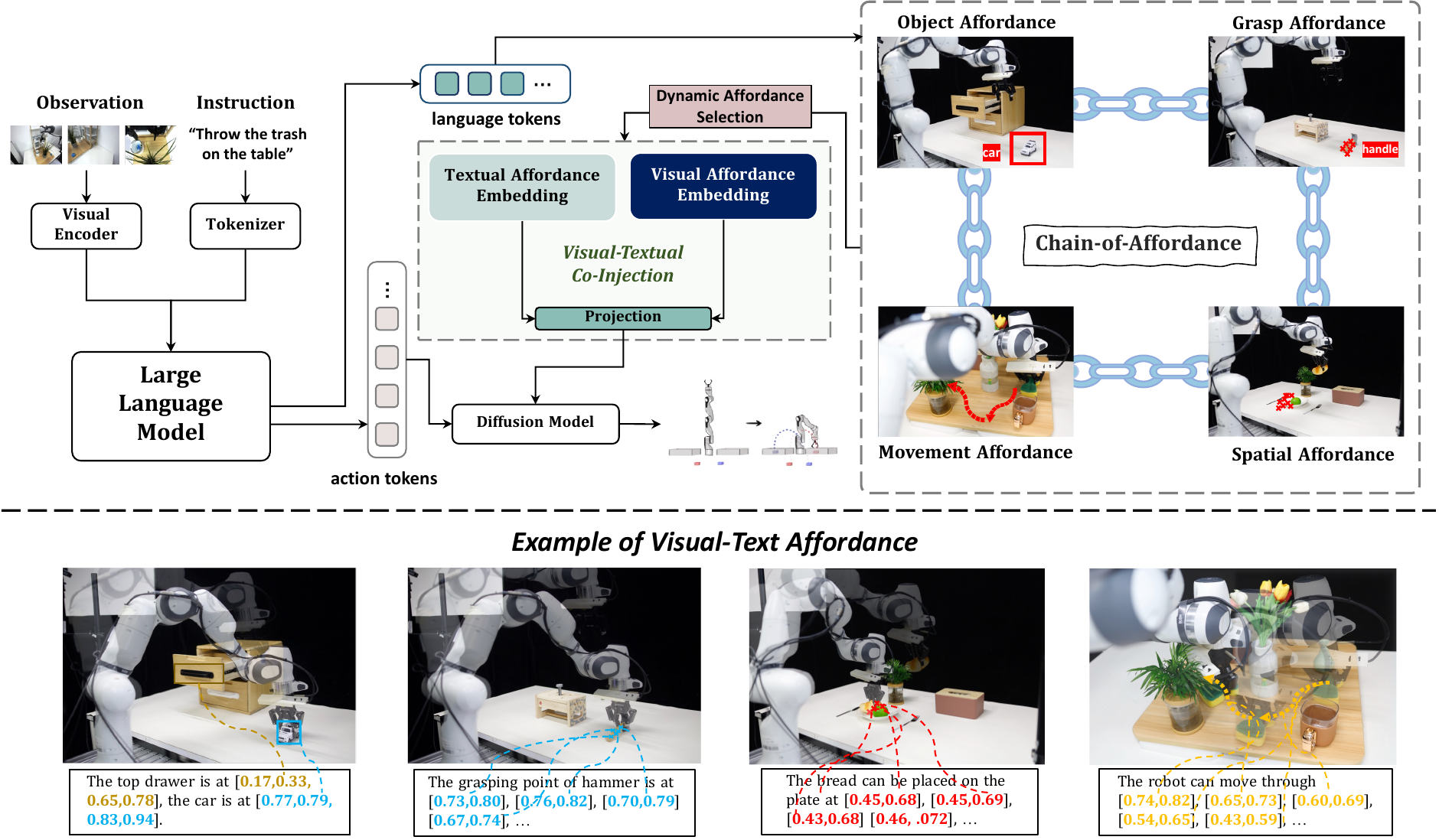}
\captionof{figure}{This figure illustrates the overall framework of our CoA-VLA model, which empowers vision-language-action models with chain-of-thought reasoning capabilities for generalizable visuomotor policy learning. We achieve this by designing four distinct types of affordance and introducing a novel visual-text co-injection method to integrate this knowledge into the decision-making process.}
\label{fig:workspace}
}
]
\input{sec/0_abstract}
\input{sec/1_intro}

\input{sec/2_related}
\input{sec/3_method}

\input{sec/4_experiments}
\input{sec/5_conclusion}
\input{sec/Acknowledegments}

{
    \small
    \bibliographystyle{ieeenat_fullname}
    \bibliography{bib_file/ntp, bib_file/vlm, bib_file/robot, bib_file/reasoning, bib_file/affordance}
}

\input{sec/X_suppl}

\end{document}

%% file: sec/0_abstract.tex
\begin{abstract}

Robot foundation models, particularly Vision-Language-Action (VLA) models, have garnered significant attention for their ability to enhance robot policy learning, greatly improving robot's generalization and robustness. OpenAI’s recent model, O1, showcased impressive capabilities in solving complex problems by utilizing extensive reasoning chains. This prompts an important question: can robot models achieve better performance in multi-task, complex environments by reviewing prior observations and then providing task-specific reasoning to guide action prediction?
In this paper, we introduce \textbf{Chain-of-Affordance  (CoA-VLA)}, a novel approach to scaling robot models by incorporating reasoning in the format of sequential robot affordances to facilitate task completion. Specifically, we prompt the model to consider the following four types of affordances before taking action: (1) \textit{object affordance} — what object to manipulate and where it is; (2) \textit{grasp affordance} — the specific object part to grasp; (3) \textit{spatial affordance} — the optimal space to place the object; and (4) \textit{movement affordance} — the collision-free path for movement. We further transform each affordance into two prompting formats: \textbf{\textit{visual affordance and textual affordance}}. We introduce a novel vision-language co-injection module that integrates this knowledge into the policy network. This allows the robot to leverage essential contextual information during action inference, resulting in improved precision and robustness. Our experiments demonstrate that CoA-VLA outperforms state-of-the-art robot foundation models, including OpenVLA and Octo, on a variety of tasks. Furthermore, CoA-VLA exhibits strong generalization capabilities, including recognizing unseen object poses, identifying free space, and avoiding obstacles in novel environments.

\end{abstract}

%% file: sec/1_intro.tex
\section{Introduction}
\label{sec:intro}

Recent advancements in Vision-Language-Action (VLA) models have shown that training with internet-scale data can empower end-to-end policy learning models to outperform non-VLA models. However, current approaches often rely heavily on high-level planning or task decomposition by off-the-shelf Large language models(LLMs) or Vision language models (VLMs), limiting models from developing implicit reasoning on their own. OpenAI's recent O1 model has demonstrated that LLMs can improve performance on complex tasks through extensive reasoning chains. If this reasoning capability can be applied to the control model, it could enhance their action robustness and generalizability. Yet, the exploration of self-driven reasoning in robotics remains under-explored, highlighting an important frontier for future research.


In this work, we propose \textbf{Chain-of-Affordance}, namely \textbf{\methodname}, a novel perspective on generalizing model reasoning at test-time, and leverage such generated reasoning to facilitate the policy learning process. Our model builds upon the DiffusionVLA~\cite{diffusionvla}, a state-of-the-art VLM model combining autoregression and diffusion objectives. Our method leverages visual affordance in robot learning, conceptualizing various actions and interactions with objects or the environment that a robot can perform based on visual contexts. We consider four types of affordance that are critical for robots to understand their observational surroundings to interact effectively with objects and achieve tasks in dynamic: 
\\
1) \textbf{Object affordance}. When a user provides vague instructions, the robot should be capable of identifying the target object for interaction and its location within the view.
\\
2) \textbf{Grasp affordance}. It involves the robot assessing the object's most appropriate points or surfaces to enable secure and stable handling.
\\
3) \textbf{Spatial affordance}. The robot needs to identify a set of coordinates that satisfy relationships described in language, such as free space for placement.
\\
4) \textbf{Movement affordance}. Identifying a trajectory for the robot to move without collision is crucial in the real world to avoid catastrophic damage. 

These four affordances form a sequential chain, requiring the robot to possess prior knowledge at each step to advance to the next. Specifically, at test time, the robot must first understand ``what to manipulate and where the object is located." Next, it determines ``how to grasp the object and finally'', ``where to place it," all while following a trajectory that ensures safe task completion. During inference, the affordance chain is progressively generated as the action state evolves, avoiding unnecessary computational costs associated with outputting extensive language. We introduce two formats for chain-of-affordance reasoning: text-based and image-based chain-of-affordance prompting. Our model's architecture natively supports text prompting integration, but image prompting requires adaptation. To address this, we developed an image affordance injection module that seamlessly integrates visual affordances into the policy network. Once generated by the model, this affordance knowledge is reused in policy learning, enabling the model to produce robust and generalizable actions.

We conduct extensive experiments on both simulated benchmarks and real-world tasks. Specifically, on the LIBERO~\cite{liu2024libero} benchmark, our proposed CoA-VLA outperforms several other state-of-the-art approaches, including the Diffusion Policy~\cite{diffusion-policy}, Octo~\cite{octo}, and OpenVLA~\cite{openvla}. Due to the substantial sim-to-real gap, many existing studies also evaluate real robots. We set up seven real-world robot tasks, including long-horizon and challenging tasks such as serving tea, cleaning garbage, and wiping tables. We perform multi-task learning on these real-world tasks and demonstrate that the VLA model, enhanced by chain-of-affordance reasoning, successfully handles most of these complex tasks with a high average success rate. Furthermore, CoA-VLA exhibits strong generalization capabilities to handle unseen object poses, adjust motion when obstacles appear, and identify free space for placement. 

The core contribution of our work lies in a novel framework that enhances vision-language-action (VLA) models by integrating affordance-aware reasoning. Central to this framework is a new module designed to strategically infuse affordance knowledge into policy learning, enabling robots to better ground actions in physical and contextual understanding. While our approach draws inspiration from prior studies on affordance, the synthesis of these concepts with VLA systems represents a significant advancement. Specifically, our framework uniquely combines these components to forge more robust, interpretable, and generalizable robotic policies, addressing critical limitations in existing VLA systems that often overlook the role of object affordances in decision-making.





%% file: sec/2_related.tex
\section{Related Works}
\label{sec:related_works}

\noindent
\textbf{Affordance in robotics.} 
In robot learning, the concept of affordance is interpreted in various ways. Typically, affordance is defined as the functions of an object, encompassing what the object is, how it can be manipulated, and its spatial relationship to the target. This concept extends beyond visual properties, linking observations directly to actions. The efficacy of affordance prediction has been shown by many learning-based manipulation methods for 6-DoF grasping~\cite{fang2023anygrasp, fang2020graspnet, Contact-graspnet} and stable object placement. It can also represented in many ways~\cite{ murali2021same, liu2022structformer, zeng2021transporter, yuan2023m2t2, jiang2021synergies, li2024manipllm, black2023zero, liu2024moka,huang2024rekep} such as part segmentation, dense image feature descriptors and keypoints. Some methods leverage human videos to obtain affordances~\cite{bahl2023affordances}, while others use vision-language models (VLMs) to predict points representing spatial affordances~\cite{chen2024spatialvlm, yuan2024robopoint}. RT-Affordance~\cite{rt-affordance} employs more descriptive affordance representations, and TraceVLA~\cite{anonymous2024tracevla} incorporates visual traces as additional input to enhance VLA. Our approach introduces multiple types of affordances and formulates them as chain-of-thought reasoning to further improve VLA model performance.
\\
\\
\noindent
\textbf{Reasoning for language and control.} 
Prompting large language models (LLMs) with "think step-by-step"~\cite{wei2022chain} has significantly advanced their ability to solve complex tasks. Numerous approaches~\cite{besta2024graph, zhou2022least} have since been developed to encourage deeper reasoning in LLMs, such as Tree-of-Thought~\cite{yao2024tree} and Chain-of-Code~\cite{li2023chain}, establishing this as a standard practice in language modeling. Recent research has leveraged LLMs and vision-language models (VLMs)~\cite{huang2023voxposer, huang2024copa, li2024manipllm, duan2024manipulate, huang2024rekep, huang2024a3vlm,huang2024rekep} as high-level planners in robotics, often using fine-tuned, open-source models or closed-source LLMs alongside policy networks for low-level task execution. These studies illustrate that detailed reasoning can enhance low-level control. ECoT~\cite{embodiedcot} introduces a reasoning strategy for VLMs that includes task decomposition, subtask descriptions, fine-grained movement instructions, gripper positioning, and object tracking on the table. CoT-VLA~\cite{cot-vla} generates sub-goa for autoregressive VLA to guide action. Unlike existing approaches, our work introduces an affordance taxonomy (categorized into four types) to unify textual and visual affordance representations in robot learning. Coupled with a dynamic selection mechanism, this taxonomy enables adaptive policy training, where task-relevant affordances are prioritized at each timestep. The resultant framework achieves computationally efficient reasoning while retaining robustness to environmental ambiguities.

%% file: sec/3_method.tex
\section{Preliminary on Vision-Language-Action Models}
Our work builds upon Vision-Language-Action (VLAs) as the backbone for our chain-of-affordance policies. VLAs employ a straightforward policy-learning approach: beginning with a pre-trained vision-language model, they fine-tune it to predict the next robot action \( a \) based on the current image observation \( I \), task instruction \( T \), and reasoning \( r \). There are two types of VLAs: autoregressive VLAs~\cite{brohan2023rt-2, o2023open-x, anonymous2024tracevla, embodiedcot, openvla,pertsch2025fast,cotvla} and diffusionVLAs~\cite{wen2024tinyvla, roboflamingo, [pi0, diffusionvla}. The former uses discrete action tokens within the vision-language model’s vocabulary, enabling action generation similar to language modeling through next-token prediction. The latter leverages a policy head~\cite{black2023zero, black2023training, lin2024datascalinglawsimitation}, such as a diffusion policy~\cite{wen2024tinyvla,li2024cogactfoundationalvisionlanguageactionmodel} or flow matching~\cite{[pi0}, to output continuous robot actions.

In this work, we employ the recently released DiVLA~\cite{diffusionvla} model, which integrates the Qwen2-VL~\cite{wang2024qwen2} vision-language model with a diffusion model head for action prediction. In the following sections, we will discuss how we improve this VLA by enabling it to reason through robot affordances before selecting an action.


\section{Methodology}
This section introduces our approach to explicitly leveraging affordance as a foundation for the robot models. In Section~\ref{sec:definition}, we formally define the concept of chain of affordance as a structured sequence of affordances, each representing an actionable insight. We provide detailed descriptions of four distinct types of affordances that together constitute this chain, explaining how each type contributes to understanding and executing complex tasks. In Section~\ref{sec:formatting}, we present two formats for representing the chain of affordances: a text format and an image format. We then discuss how these representations can be integrated into the policy learning process. Finally, in Section~\ref{sec:generating_data}, we outline the pipeline used to automatically generate large-scale chain-of-affordance data.



\begin{figure*}[t]
    \centering
    \includegraphics[width=\textwidth]{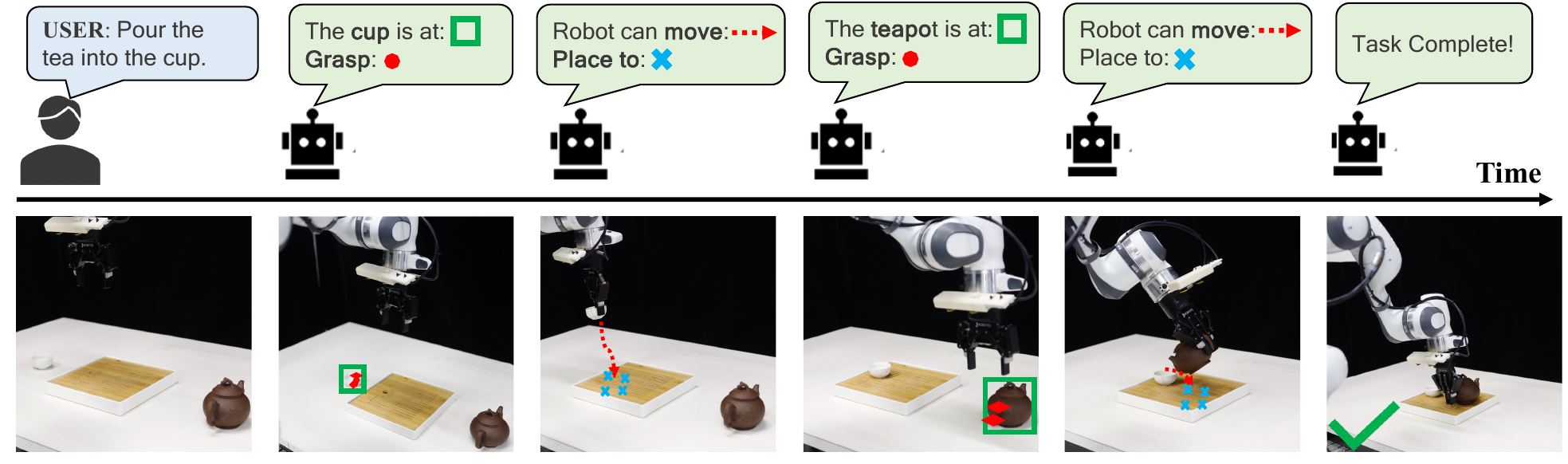}
    \caption{\textbf{An example of the chain-of-affordance for the PourTea task.} The first row presents the text affordance and the second row shows the visual affordance. By employing a dynamic affordance selection mechanism, our method avoids generating redundant affordances at every timestep.}\label{fig:example_coa}
\end{figure*}
\subsection{Definition of Chain-of-Affordance}
\label{sec:definition}
Given a dataset $\mathcal{D} = \{(\tau_1, g_1), \dots, (\tau_N, g_N)\}$ consisting of $N$ expert demonstrations, where each demonstration $\tau_i$ is paired with a task description $g_i$ in natural language. Each task description $g \in \mathcal{G}$ specifies a composition of multiple sub-tasks, and each demonstration $\tau_i$ is represented by a sequence of observations. We define $z \in \mathcal{Z}$ as the affordance-based reasoning in natural language that guides the task. The model decomposes $z$ into four components, $z = \{z_{obj}, z_{grasp}, z_{spat}, z_{move}\}$, where $z_{obj}, z_{grasp}, z_{spat},$ and $z_{move}$ represent object, grasp, spatial, and movement affordances, respectively. Our objective is to learn an intermediate language output $z: \mathcal{O} \times \mathcal{G} \rightarrow \mathcal{Z}$ that maps observations and task descriptions to affordance reasoning in natural language. This intermediate output provides specific guidance for action generation, enabling the generation of low-level actions $a \in \mathcal{A}$. Note that low-level actions are generated conditioned on the demonstration, task description, and affordance reasoning: $a \sim p(a|\tau, g, z)$.


\textbf{Details for diverse robot affordance.}
In our approach, we model the robot affordance in the format of natural language as intermediate outputs. We will give detailed description on each type of affordance. 

\textbf{\textit{Object affordance:}} Object affordance equips robots with the foundational ability to autonomously determine which object to interact with and where it is located, particularly in scenarios where user instructions lack explicit spatial or semantic details. This capability bridges the gap between ambiguous queries (e.g., ``Pour the drink") and actionable execution by enabling the robot to: 1) Identify the target object through natural language grounding (e.g., resolving ``drink" to ``teapot"), 2) Localize the object within its environment using pixel-aligned bounding box predictions. In our framework, object affordance is operationalized through two tightly coupled components: 1) \textbf{Semantic identification}: Resolving object names from free-form language input, and 2) \textbf{Spatial grounding}: Predicting the object’s 2D location via visual scene understanding. By integrating these capabilities, the robot establishes a contextual foundation for downstream decision-making, ensuring interactions are both intention-aware (aligned with user goals) and environmentally grounded (physically feasible).



\textbf{\textit{Grasp affordance:}} Grasp affordance encompasses the possible functions or ways an object can be manipulated. This affordance goes beyond visual characteristics, linking observations directly to actions, and is crucial for tasks requiring 6-DoF (degrees of freedom) grasping~\cite{jiang2021synergies, sundermeyer2021contact, murali2021same}. Prior work has demonstrated the effectiveness of affordance prediction for stable object handling and placement. Representations of grasp affordance vary, including part segmentation or keypoints. In our work, we use a set of 2D points to represent the grasping point for an object. 

\textbf{\textit{Spatial affordance:}} Spatial affordance centers on a robot’s ability to interpret and reason about spatial relationships within 3D environments, enabling tasks such as identifying collision-free regions for object placement or navigation. For instance, RoboPoint~\cite{yuan2024robopoint} locates free space, and SpatialVLM~\cite{chen2024spatialvlm} predicts spatial relations quantitatively and qualitatively. In our framework, spatial affordance is operationalized as actionable destinations—discrete 2D coordinates representing feasible interaction zones. 

\textbf{\textit{Movement affordance:}} Movement affordance defines the trajectory a robot can follow during a task. This path may change depending on environmental factors, such as obstacles introduced along an intended trajectory. By modeling movement affordance, we provide the robot with adaptable paths for action, allowing it to respond dynamically to environmental changes and complete its task effectively. These affordances collectively enable the robot to understand and act upon various elements within its operational space, enhancing its interaction capabilities and responsiveness.

\textbf{Dynamic affordance selection.} Our proposed method formulates affordances as combined text and visual prompts.  While this approach offers benefits, it introduces additional computation at test time, potentially slowing down the algorithm. We observe that it's unnecessary to utilize all available affordances during testing. For example, once an object is picked up, predicting its object affordance and the grasp affordance becomes redundant. The specific affordances required in a given sub-step depend on the task's progress and the environment. Therefore, we implement dynamic affordance selection, adaptively choosing the necessary affordances at both training and test times to reduce computational cost.
Several methods can achieve this, such as gradient-based selection~\cite{xu2014gradient}. Our approach prioritizes simplicity by leveraging proprioception. Proprioception refers to information about the robot's state, including joint angles and other movement data. We transform this proprioceptive data into a single token and concatenate it with the visual token before feeding it into the large language model. Training on large-scale annotated datasets like Droid~\cite{khazatsky2024droid} enables our model to intelligently select relevant affordances at each time step. This is straightforward for the model to learn. For instance, if the proprioceptive state indicates a partially closed gripper and the wrist-mounted camera detects an object, the model can infer that the object and grasp affordances are likely unnecessary. Instead, it can focus on the movement affordance to guide the action trajectory and the spatial affordance to determine a suitable placement location. We found this strategy to be simple and useful, reducing the model's computational cost without hurting performance.

\begin{figure*}[t]
    \centering
    \includegraphics[width=0.73\textwidth]{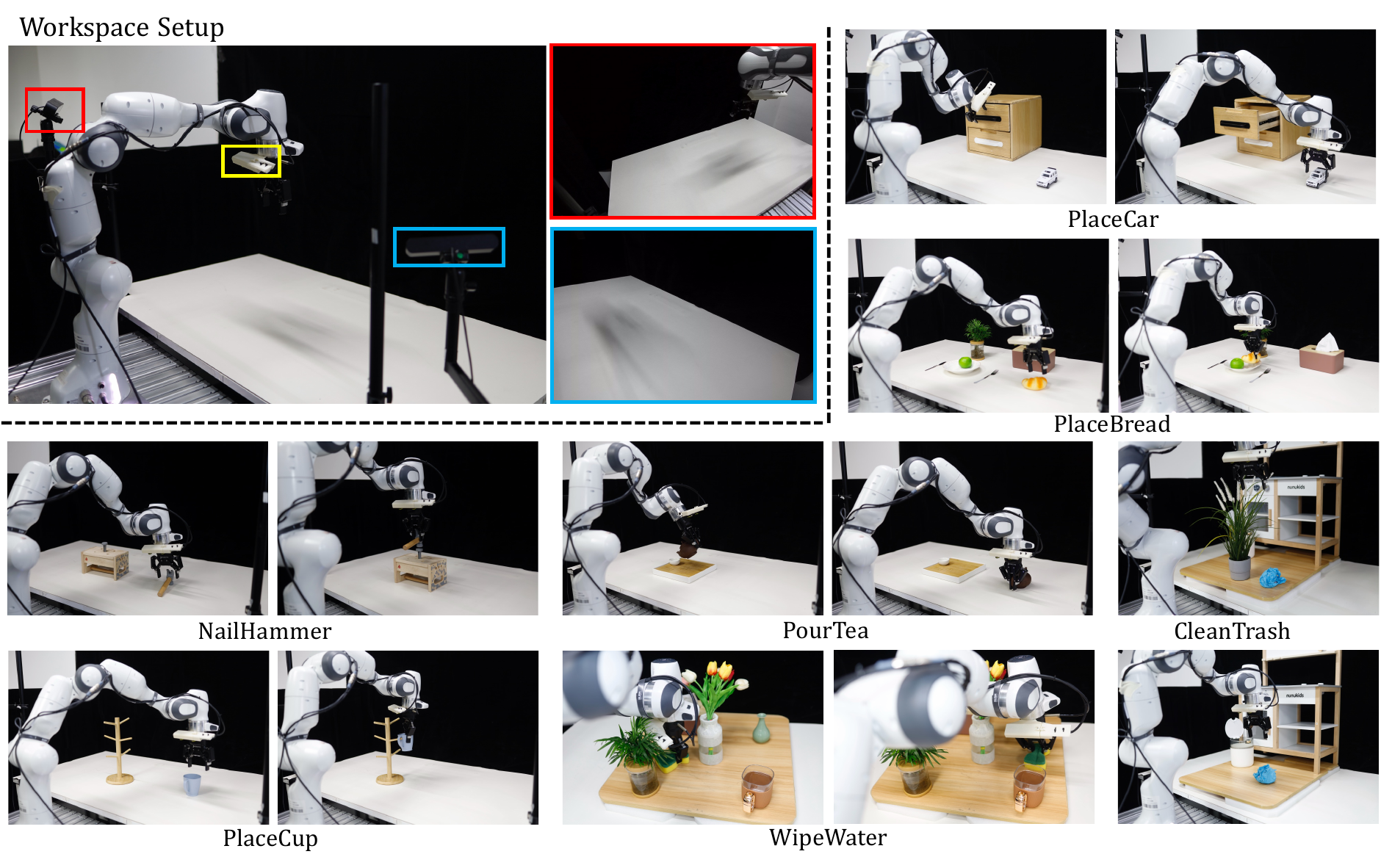}
    \caption{\textbf{Robot setup and examples for real-world manipulation tasks.} We evaluate seven real-world tasks on Franka robot arm equipped with two external Zed cameras and a Realsense 435i wrist camera.}\label{fig:robot_setup}
\end{figure*}

\subsection{Formatting of Visual-Textual Chain-of-Affordance}
\label{sec:formatting}
In this section, we present a multimodal chain-of-affordance framework that formalizes affordance reasoning through two complementary representations: (1) \textbf{textual affordance}, which provides a structured, language-driven representation of actionable possibilities (e.g., ``graspable handle" or ``navigable path"), and (2) \textbf{visual affordance}, which delivers a spatially grounded, scene-aware perspective via pixel-aligned cues. While the textual format enables explicit semantic reasoning, the visual modality enhances interpretability in visually dense or ambiguous environments by grounding affordances in observable scene geometry.

To unify these modalities, we introduce a visual-language co-injection module that dynamically aligns and integrates visual affordance prompts in a shared parameter space. This module bridges the gap between abstract language-based reasoning and pixel-level visual context, enabling the policy model to synergistically leverage both modalities for robust, context-aware action generation.

\textbf{Textual affordance.} Natural language serves as the predominant modality for representing visual affordances due to its semantic richness and compatibility with human-AI interaction. For instance, object affordances can be encoded as coordinate pairs (e.g., bounding boxes localizing a "graspable cup"), while spatial affordances are expressed as coordinate pairs defining navigable regions (e.g., "reachable shelf area"). Figure~\ref{fig:example_coa} (top) illustrates how diverse affordances (e.g., "movable," "stackable") are mapped to natural language descriptors.To avoid rigid linguistic templates that might bias policy learning, we employ ChatGPT to dynamically paraphrase affordance descriptions with varied syntactic structures and vocabulary (e.g., alternating between "placeable surface" and "flat area for stacking"). This strategy, analogous to data augmentation in NLP, enhances the model’s robustness to linguistic variability while preserving its conversational fluency.

\textbf{Visual affordance.} Inspired by visual prompting methods that enhance model interpretability, we propose visual affordance augmentation—a technique that augments natural language affordance descriptions with direct, pixel-aligned visual cues. This approach encodes affordances by overlaying coordinate markers (e.g., bounding boxes, interaction points) or motion trajectories onto the robot’s historical observation frames (Figure~\ref{fig:example_coa}, bottom). These annotations act as chain-of-affordance prompts, visually grounding the model’s understanding of actionable properties like "graspable" or "navigable" within the scene’s geometry. By embedding affordances directly into the visual input, we create an explicit structure that bridges the gap between abstract language and actionable visual context.For movement affordances, we employ thin, low-saliency trajectories to avoid overshadowing critical scene elements while guiding motion planning. Conversely, key interaction zones — such as grasping points, object bounding boxes, and spatial affordances — are rendered with thicker, high-contrast overlays (e.g., semi-transparent colors) to ensure visual salience without occluding environmental context. This hierarchical visual encoding distinguishes affordance types at a glance, enabling the model to prioritize task-relevant cues while maintaining computational efficiency.

\textbf{Visual-textual co-injection.} To integrate visual and textual affordances into action generation, we design a unified embedding pipeline that seamlessly fuses both modalities into the diffusion policy. For textual affordances, we use the last embedding from the VLM models and add an MLP layer to tokenize it. Concurrently, visual affordances are processed into patch tokens via a pretrained Vision Transformer (ViT-Small). 
The fused tokens are then processed by two standard Transformer blocks, balancing computational efficiency with sufficient expressive power for cross-modal reasoning. The encoder’s output embeddings are projected into the diffusion model using FiLM conditioning layers, inspired by MT-ACT~\cite{roboagent}. This conditioning dynamically modulates the diffusion process by injecting affordance-aware features, enabling the policy to generate actions that respect both spatial constraints and semantic intent.
By unifying affordance modalities early in the pipeline, our method retains the computational efficiency of standard diffusion frameworks while enhancing robustness. The FiLM-based conditioning acts as a bottleneck, distilling only the most salient affordance cues to guide action generation, thereby avoiding overfitting to redundant visual or linguistic signals.

\input{sec/table/franka_exp}
\subsection{Generating Chain-of-Affordance Data}
\label{sec:generating_data}
To prevent overfitting on affordance diversity, generating large-scale, high-quality data is crucial. While the standard approach typically relies on direct human labeling, this method is both costly and labor-intensive. Therefore, in this section, we outline a comprehensive pipeline for automatically generating diverse affordances using a range of tools to streamline the data generation process.

Our pipeline begins with GPT-4o~\cite{chatgpt}, which generates a detailed description of the scene and identifies relevant entities from the language instructions. This allows us to create contextually rich, entity-specific affordances tailored to the task at hand. Using these entities, we leverage Grounding DINOv2~\cite{liu2303marrying} and SAM~\cite{sam, ravi2024sam} to produce bounding boxes around each identified object within the scene. SAM initially provides masks, which we convert into bounding boxes, and we finalize these by taking out the intersection over union (IoU) between the outputs of Grounding DINOv2 and SAM. This IoU-based refinement ensures that bounding boxes are accurately aligned with object contours. At this stage, we also capture the gripper’s position for subsequent affordance calculations.To represent spatial affordance, we integrate RoboPoint~\cite{yuan2024robopoint}, a state-of-the-art model that predicts spatial affordances directly within the image. Additionally, we prompt GPT-4o to annotate spatial points based on the scene context. The spatial predictions from RoboPoint and GPT-4o are then combined, after which we cluster these points to form a coherent representation, eliminating any outliers to maintain accuracy. This process captures spatial affordances that are not only visually aligned but contextually relevant.For capturing movement trajectories, we employ CoTracker~\cite{karaev2023cotracker, karaev2024cotracker3}, an advanced transformer-based tracking model. CoTracker enables us to follow the robot gripper's path, recording its movement through the scene and gathering essential trajectory data. This movement data provides insight into how the robot interacts dynamically with the environment, adding temporal dimensions to our affordance representation. The combination of spatial affordances, grasping points, and movement trajectories enables us to model a rich, multi-faceted affordance landscape for each scenario. This automated, tool-assisted pipeline produces a detailed and diverse set of affordance annotations, significantly reducing the need for manual labeling.


%% file: sec/table/franka_exp.tex
\begin{table*}[t]
\centering
\vspace{0.1cm}
\caption{\textbf{Experimental results for multi-task learning.} 
Our method achieved the best performance in both the in-distribution test setup and under visual changes.}
\label{tab:main_result}
\resizebox{\textwidth}{!}{\begin{tabular}{c|ccccccc|c}
\toprule
  &   \multicolumn{7}{c}{Seven Tasks on Franka Robot Arm}   \\
\midrule
\multirow{2}{*}{\textbf{Model $\setminus$ Tasks}}  & \multicolumn{7}{c}{In-Distribution} & \\
   &CleanTrash  & PourTea  & NailHammer & PlaceBread & PlaceCar & WipeWater& HangCup & Avg.\\
\midrule
Diffusion Policy~\cite{diffusion-policy}  &4/11 &0 &8/11 &3/11&9/11&2/11 & 7/11  & 33/77 (42.93\%)\\ 
Octo~\cite{octo} & 4/11 & 0/11 & 7/11 &4/11& 8/11&3/11 & 8/11 & 34/77 (44.13\%) \\
OpenVLA~\cite{openvla} &4/11 & 0/11 & 9/11& 6/11 & 9/11& 5/11 &9/11 & 52/77 (54.89\%)\\
DiffusionVLA~\cite{diffusionvla} &\colorbox{AliceBlue}{\textbf{10/11}} &6/11 &\colorbox{AliceBlue}{\textbf{10/11}} & 7/11 &\colorbox{AliceBlue}{\textbf{10/11}} & 6/11 &\colorbox{AliceBlue}{\textbf{10/11}} & 59/77 (76.60\%)\\
\textbf{\methodname} & \colorbox{AliceBlue}{\textbf{10/11}} & \colorbox{AliceBlue}{\textbf{8/11}} & 9/11 & \colorbox{AliceBlue}{\textbf{10/11}} & \colorbox{AliceBlue}{\textbf{10/11}} & \colorbox{AliceBlue}{\textbf{9/11}} & \colorbox{AliceBlue}{\textbf{10/11}} & \colorbox{AliceBlue}{\textbf{64/77 (85.54\%)}} \\ 
\midrule
 \multirow{2}{*}{\textbf{Model $\setminus$ Tasks}} & \multicolumn{7}{c}{Visual Generlization} &\\
   &CleanTrash  & PourTea  & NailHammer & PlaceBread & PlaceCar & WipeWater& HangCup & Avg.\\
 \midrule
 Diffusion Policy~\cite{diffusion-policy} & 0/9&0/9 &0/9&0/9&  3/9 &0/9 &0/9& 3/63 (4.76\%)\\
 Octo~\cite{octo} &0/9 &0/9& 3/9 &0/9 &4/9& 2/9& 3/9 & 12/63 (19.05\%)\\
 OpenVLA~\cite{openvla} & 3/9 &0/9 & 0/9 & 4/9 &4/9&3/9& 0/9 & 14/63 (22.22\%) \\
DiffusionVLA~\cite{diffusionvla} &  2/9 & 2/9 & 5/9 & 5/9& \colorbox{AliceBlue}{\textbf{5/9}} & 3/9 & 6/9& 28/63 (44.44\%)\\
\textbf{\methodname} & 
\colorbox{AliceBlue}{\textbf{4/9}} & \colorbox{AliceBlue}{\textbf{4/9}} & \colorbox{AliceBlue}{\textbf{6/9}} & \colorbox{AliceBlue}{\textbf{6/9}} & \colorbox{AliceBlue}{\textbf{5/9}} & \colorbox{AliceBlue}{\textbf{4/9}} & \colorbox{AliceBlue}{\textbf{7/9}}&\colorbox{AliceBlue}{\textbf{36/63 (57.14\%)}}\\ 
\bottomrule
\end{tabular}}
\vspace{-0.5cm}
\end{table*}

%% file: sec/4_experiments.tex
\input{sec/table/libero}

\section{Experiments}
The objective of our experimental evaluations is to assess the efficacy of CoA-VLA as a reasoning framework that augments the policy-learning capabilities of a baseline robot foundation model. Specifically, we seek to address the following questions:
1) Does the VLA model demonstrate improved performance with the integration of CoA in real-world experiments?
2) What is the performance of CoA-VLA in simulation benchmarks?
3) How effectively does CoA-VLA generalize to challenging scenarios? 4) How important is our proposed visual-textual affordance approach, compared to the vanilla VLA? We will give a detailed analysis in this section. Note that some experiments are presented in the Appendix due to limited space.

\subsection{Evaluation on Real Robot}
\textbf{Experimental setup.} We conduct evaluations using the 6-DoF Franka robot arm, a widely adopted setup for assessing generalizable robot policies. Our setup includes two third-person cameras (ZED cameras) positioned on either side of the robot arm, as well as an egocentric camera (Realsense D435i) mounted on the wrist. We design seven challenging tasks, such as pouring tea and cleaning the table, incorporating both short-horizon and long-horizon tasks to test the adaptability of our approach. We illustrate our robot setup and tasks with examples in Figure~\ref{fig:robot_setup}. Detailed descriptions of each task and the experimental setup, and our ablation experiments  are provided in the Appendix. 

\textbf{Train data.}
We use the Droid dataset~\cite{khazatsky2024droid} as an external data source, filtering out samples without language annotations, leaving 39K trajectories. Our pipeline generates synthetic chain-of-affordance data for pre-training. The model is then post-trained on 692 trajectories across seven tasks.

\textbf{Baseline.} We compare CoA-VLA with Diffusion Policy~\cite{diffusion-policy}, Octo~\cite{octo}, OpenVLA~\cite{openvla}, TinyVLA~\cite{wen2024tinyvla}, DiffusionVLA~\cite{diffusionvla}. The latter three methods are all vision-language-action models that achieve state-of-the-art performance in the real world. Notably, the DiffusionVLA is the same model our approach is built upon, but trained with vanilla reasoning instead of our proposed chain-of-affordance. To ensure a fair comparison, all models are fine-tuned on the same dataset we use for training our approach. All models are trained with the same number of iterations, and the last checkpoint is used for evaluation. It ensures that no model is cherry-picked for comparison. 

\textbf{Real robot experimental results.} 
The experimental results are presented in Table~\ref{tab:main_result}. We assess the performance using two different settings: the in-distribution setup and visual generalization testing. In the in-distribution setup, our method surpasses all SOTA robot foundation models in terms of average success rate. Notably, our method exceeds the performance of OpenVLA by 30.65\%, despite having a significantly smaller model size and less pre-training data. Compared to our baseline model, which employs vanilla reasoning, our method achieves a 14.29\% increase in accuracy. To assess visual generalization, we further evaluate these advanced models under varying visual conditions. As observations become more complex—such as through the addition of distractors or vibrant backgrounds—the performance gap between CoA-VLA and OpenVLA/DiffusionVLA widens. These findings highlight the critical role of embodied, task-specific chain-of-affordance reasoning in enhancing the performance of VLA models.


\subsection{Evaluation on Simulation}
In this section, we examine the performance of CoA-VLA on LIBERO~\cite{liu2024libero}. LIBERO is a robot learning benchmark comprising over 130 language-conditioned manipulation tasks. We follow the setting as in OpenVLA~\cite{openvla} open-sourced code and test on four task suites: LIBERO-Spatial, LIBERO-Goal, LIBERO-Object, and LIBERO-Long. The detailed experimental setup is in the Appendix.

\noindent
\textbf{Simulation experimental results.} We compare with the Diffusion Policy~\cite{diffusion-policy}, ScaleDP~\cite{zhu2024scalingdp}, Octo~\cite{octo}, and OpenVLA~\cite{openvla}. The experimental results of LIBERO are presented in Table~\ref{tab:libero_comparison}. Our findings indicate that \methodname~consistently achieves superior performance across all evaluated settings, securing the highest success rate among the methods tested. Specifically, \methodname~achieves an overall success rate of 79.8\%, outperforming OpenVLA, the previous best-performing method, by a margin of 3.3\%. This improvement demonstrates the effectiveness of our approach in simulation. Compared to pre-trained robotic models like Octo and OpenVLA, \methodname~demonstrates non-trivial gains, showing its effectiveness in leveraging prior knowledge while adapting to diverse task requirements. Additionally, when benchmarked against train-from-scratch methods such as Diffusion Policy and ScaleDP, \methodname~exhibits substantial improvements, underscoring the advantages of our approach.

\begin{figure}[t]
    \centering    \includegraphics[width=0.42\textwidth]{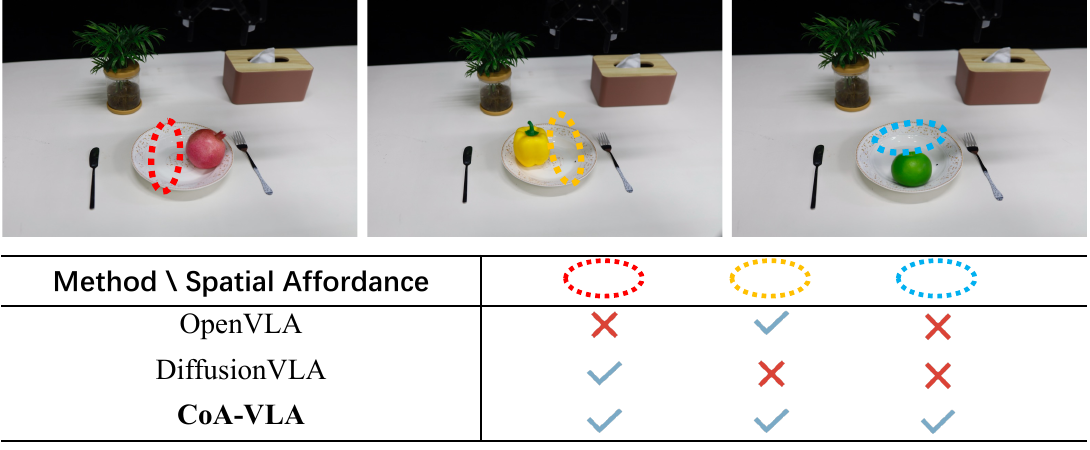}
    \caption{\textbf{Spatial affordance for \methodname.} \methodname~ can identify free space for object placement..}\label{fig:spatial_gen}
\end{figure}
\begin{figure}[t]
    \centering
    \includegraphics[width=0.42\textwidth]{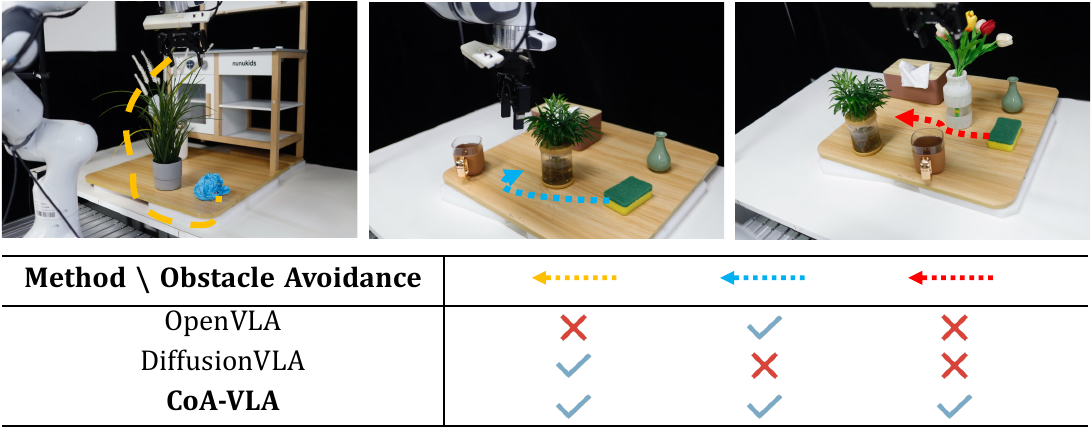}
    \caption{\textbf{Movement generalization for \methodname.} \methodname~can avoid obstacles and operate safely.}\label{fig:movement_gen}
\end{figure}
\subsection{More Experiments}
\textbf{Spatial affordance enables safe placement.} We evaluate the effectiveness of our spatial affordance approach in the PlaceBread task, as illustrated in Figure~\ref{fig:spatial_gen}. In this task, the robot is presented with a plate on which three distinct objects are already placed, and it is instructed to add a piece of bread onto the plate. Our method successfully identifies open areas on the plate, allowing it to accurately position the bread without interference, thereby enabling \methodname~to complete all three task scenarios with precision. In contrast, alternative methods, such as OpenVLA and DiffusionVLA, succeed in only one scenario each, failing to generalize across all spatial configurations. This experiment underscores the crucial role of spatial affordance in enhancing the model’s ability to recognize and utilize available space, ultimately improving task completion accuracy and reliability across varied setups.

\textbf{Obstacle avoidance.} Collision avoidance is essential for safe and effective physical interactions, as improper maneuvers can lead to significant damage or even catastrophic outcomes. We evaluated \methodname's obstacle avoidance capabilities in two specific tasks, illustrated in Figure~\ref{fig:movement_gen}. In the first task, we positioned a vase near the center of the task area to observe whether {\methodname}  could successfully navigate around the vase to retrieve a piece of paper trash. In the second task, we introduced a series of obstacles on a table, rearranging them in different configurations to assess the robot’s adaptability in maneuvering through varied layouts to complete the task. Our approach successfully completed all three scenarios, demonstrating robust collision avoidance and spatial adaptability. In contrast, OpenVLA failed to complete any of the tasks, while DiVLA succeeded in only one scenario.  These results highlight the critical role of integrating movement affordance into the model’s reasoning process, enhancing its ability to navigate complex environments and complete tasks with precision and safety.




%% file: sec/table/libero.tex
\begin{table*}[t]
    \centering
    \caption{\textbf{Experimental results for LIBERO benchmark.} We report the success rate and standard error for four task suites.}
    \label{tab:libero_comparison}
    \resizebox{\textwidth}{!}{%
    \begin{tabular}{l|c|c|c|c|c}
        \toprule
         & LIBERO-Spatial & LIBERO-Object & LIBERO-Goal & LIBERO-Long & Average \\
        Method / Task & Success Rate ($\uparrow$) & Success Rate ($\uparrow$) & Success Rate ($\uparrow$) & Success Rate ($\uparrow$) & Success Rate ($\uparrow$) \\
        \midrule
        Diffusion Policy~\cite{diffusion-policy} & $78.3 \pm 1.1\%$ & $92.5 \pm 0.7\%$ & $68.3 \pm 1.2\%$ & $50.5 \pm 1.3\%$ & $72.4 \pm 0.7\%$ \\
        ScaleDP~\cite{zhu2024scalingdp} & $79.1 \pm 0.7\%$ & $90.4 \pm 0.9\%$ & $73.6 \pm 0.8\%$ & $48.4 \pm 1.2\% $ & $72.9 \pm 0.5\%$\\ 
        Octo~\cite{octo} & $78.9 \pm 1.0\%$ & $85.7 \pm 0.9\%$ & $84.6 \pm 0.9\%$ & $51.1 \pm 1.3\%$ & $75.1 \pm 0.6\%$ \\
        OpenVLA~\cite{openvla} & $84.7 \pm 0.9\%$ & $88.4 \pm 0.8\%$ & $79.2 \pm 1.0\%$ & $53.7 \pm 1.3\%$ & $76.5 \pm 0.6\%$ \\
        \midrule
        \textbf{CoA-VLA} & \textbf{$85.3 \pm 0.9\%$} & \textbf{$93.1 \pm 1.0\%$} & \textbf{$85.8 \pm 0.9\%$} & \textbf{$55.0 \pm 1.2\%$} & \textbf{$79.8 \pm 0.5\%$}\\
        \bottomrule
    \end{tabular}%
    }
    
\end{table*}

%% file: sec/5_conclusion.tex
\section{Conclusion}
Explicit reasoning is essential for language models to handle complex tasks. In this work, we design a reasoning-aware foundation model for robotics, focusing on various affordances: object, grasp, spatial, and movement. These affordances form an interdependent chain: the robot identifies the target object and location, determines how to grasp it, decides where to place it, and navigates accordingly. By structuring this chain of affordances as intermediate language and image outputs and feeding this reasoning into the policy model, our Chain-of-Affordances (CoA-VLA) model outperforms baselines on real-world robotic tasks.CoA-VLA models also generalize well to complex environments, tackling challenges like grasping in unfamiliar orientations, avoiding obstacles, and spatial generalization. 
Our approach provides a novel perspective on designing reasoning chains to enhance embodied control.

%% file: sec/Acknowledegments.tex
\section*{Acknowledegments}
This work is supported by the National Science Foundation of China (12471501), and  the Sci-Tech Innovation Initiative by the Science and Technology Commission of Shanghai Municipality (24ZR1419000).


%% file: sec/X_suppl.tex
\clearpage
\setcounter{page}{1}
\maketitlesupplementary

\subsection{Video Demo}
We provide a video recording in the supplementary material.

\input{sec/table/task_summary}
\begin{figure*}[ht]
    \centering
    \includegraphics[width=1\textwidth]{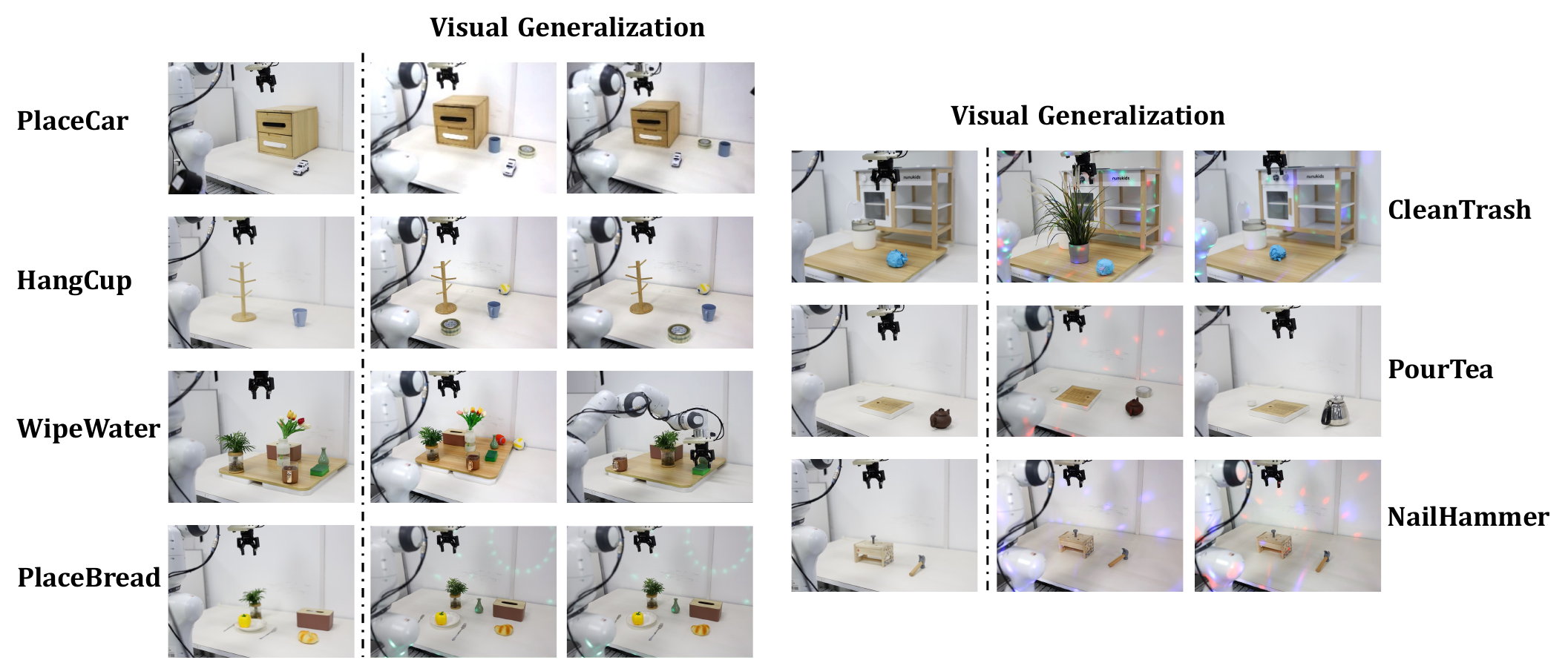}
    \caption{\textbf{Visual Generalization.} We evaluate each method on multi-task learning and visual generalization, which includes handling additional distractors and interference from colored light. We also test the ability to grasp objects of the same type but with varying shapes, such as different teapots, as well as teapots in different orientations.}
    \label{fig:visual_generalization}
\end{figure*}

\begin{figure*}[ht]
    \centering
   \includegraphics[width=0.8\textwidth]{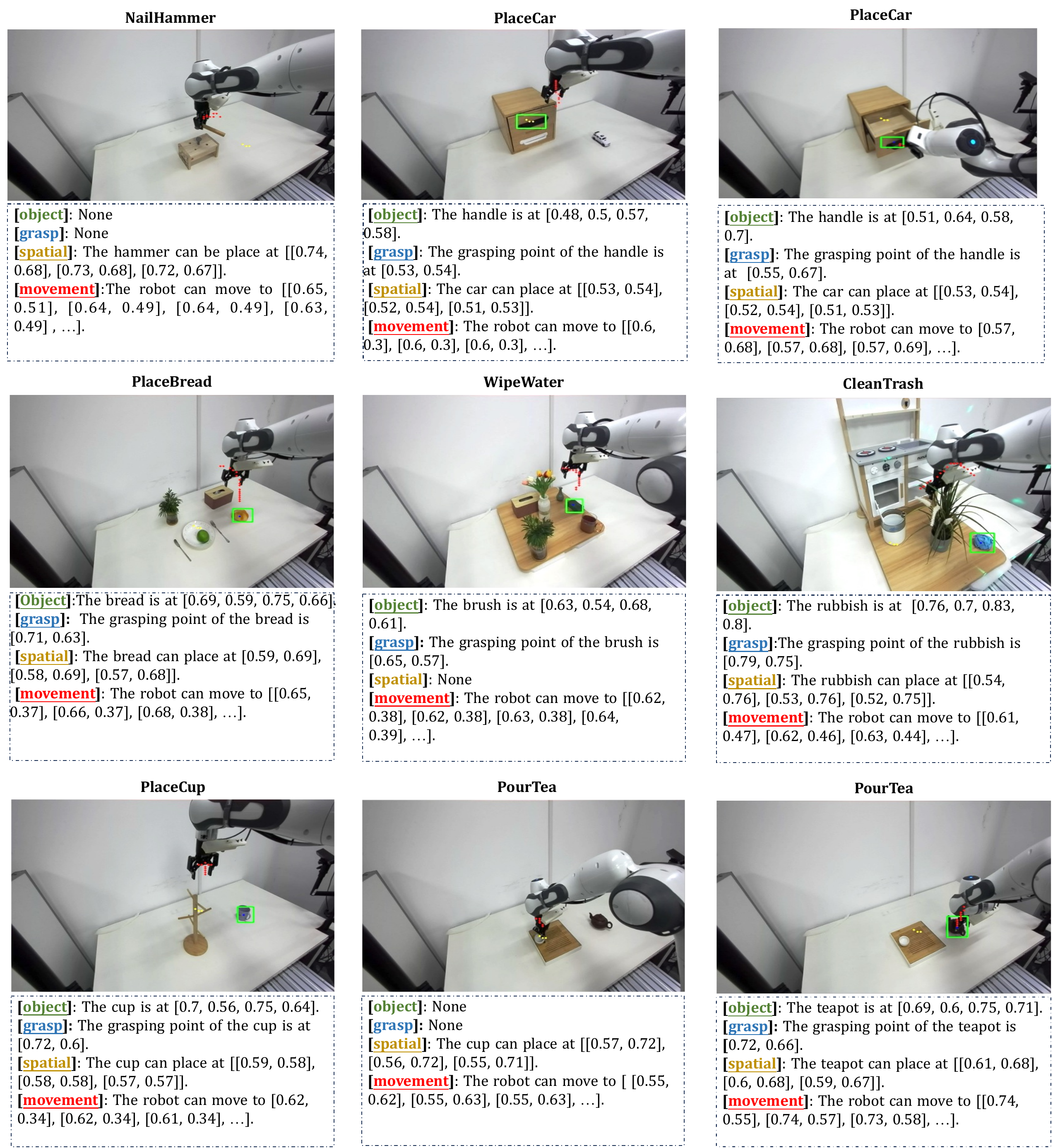}
    \caption{More detailed examples of successful Chain-of-Affordance.}
    \label{fig:example_coa_more}
\end{figure*}

\subsection{Evaluation Tasks}
In this section, we give a detailed description of the evaluated tasks that we discussed in Table~\ref{tab:main_result}. We provide the number of demonstrations for each task and the average trajectory length in Table ~\ref{tab:task-sum}.

\begin{itemize}
    \item \textbf{PlaceCar.} We randomly place the toy car on the right side of the drawer. The model is asked to pick up the toy car, put it into the drawer, and eventually close the drawer. This is a long-horizon task that requires multiple steps of action. 
    \item \textbf{PlaceBread.} The model needs to pick up the bread and place it on an empty spot on the plate, avoiding placing it on the fruit. The bread is randomly placed on the table. The model needs to pick up the bread and place it on the empty spot on the plate. 
    \item \textbf{NailHammer.} We evaluate the model's proficiency in utilizing tools effectively by assessing its ability to perform a sequence of precise actions with a hammer. The model must first identify the correct grasp point on the hammer, ensuring a stable and ergonomic grip suitable for controlled operation. It must then carefully pick up the hammer without causing it to topple or disturb its surroundings. Once the hammer is securely held, the model is tasked with driving a nail into a designated spot with precision. 
    \item \textbf{PourTea.} In this task, the robot is required to perform a sequence of actions involving a tea cup and a teapot. First, the robot must place the tea cup onto the tea tray. Next, it needs to pick up the teapot and pour tea into the teacup. Both the tea cup and the teapot are randomly positioned within a defined range on the table. A key aspect of the task is the robot's ability to accurately grasp the teapot by its stem. To ensure consistency during data collection, the tea pot's stem is always oriented facing the robot, simplifying the grasping process while still challenging the model's precision and manipulation skills.
    \item \textbf{CleanTrash.} In this task, the robot is required to perform a sequence of actions to clean up trash on a table. The task has two distinct scenarios. In the first scenario, with no obstacles, the robot must identify and pick up the randomly placed trash, then deposit it into the trash bin. The trash items are distributed across the table in a random manner. In the second scenario, a flower pot is placed on the table as an obstacle. The robot must avoid colliding with the flower pot while picking up the trash and placing it into the trash bin. The trash's location remains random, and the robot must navigate carefully to avoid knocking over the flower pot during the cleanup process. A key aspect of this task is the robot's ability to accurately avoid the flower pot while maintaining efficiency in picking up and discarding the trash.
    \item \textbf{WiperWater} In this task, the robot is required to clean up water from a table by using a sponge. The sponge is placed on the right side of the table, and the robot must pick it up and use it to wipe the water from the surface, moving from right to left. During this process, the robot must avoid any objects placed on the table, such as vases, cups, boxes, and other items. A key challenge in this task is the robot's ability to manipulate the sponge effectively while navigating around the obstacles without causing any collisions, ensuring that the entire table is cleaned efficiently. The robot's precision in both grasping the sponge and avoiding the table items is critical for completing the task successfully.
    \item \textbf{HangCup} In this task, the robot is required to pick up cups that are randomly scattered on the table and hang them on a cup rack. The robot must handle the cups carefully to avoid damaging them and ensure that the rack is not disturbed or knocked over during the process. The task challenges the robot's precision in both grasping the cups and placing them securely on the rack while maintaining stability in the environment. Successful completion relies on careful manipulation and accurate placement.
\end{itemize}

\textbf{Setup for visual generalization.} In this scenario, we evaluate the model's robustness and its ability to generalize visual perception across diverse and challenging environmental conditions. The robot is tasked with performing manipulation tasks while navigating visual complexities such as randomly placed distractors, varying lighting conditions, and a visually cluttered, colorful background. These challenges are designed to test the model's capability to stay focused on the primary task, effectively filter out irrelevant visual distractions, and adapt to dynamic and unpredictable visual environments. The objective is to ensure the robot can consistently and accurately identify and interact with target objects, even under significant deviations from typical operational settings.

\subsection{Details for Real Robot Experiments}
We train our method in a multi-task setting without relying on pre-trained weights from DiffusionVLA. Instead, we leverage our constructed dataset for pre-training. Specifically, we initialize the learning rate at 2e-5 and maintain a fixed learning rate throughout the pre-training phase, which spans 5 epochs. During this stage, the parameters of the pre-trained Vision-Language Model (VLM) are frozen, and LoRA is employed to fine-tune the model. For fine-tuning, we adopt a similar approach, starting with an initial learning rate of 2e-6. However, in this phase, we apply a cosine learning rate decay schedule and train the model for an additional 5 epochs. This training strategy ensures both effective adaptation and stability across pre-training and fine-tuning stages, optimizing the model for multi-task performance. 

For the baselines, we generally adopt a consistent training strategy. In the case of OpenVLA, the vanilla implementation utilizes only a single camera view. To extend this, we incorporate all three camera views, feeding each view into the same visual encoder and concatenating their outputs for processing. We leverage OpenVLA's pre-trained weights and trains for 20 epochs, as we observe that it typically requires a longer training time to achieve convergence. For the Diffusion Policy, we utilize DistilBERT to process language instructions, following an approach similar to YAY~\cite{yell_at_your_robot}. As for DiffusionVLA, we employ their pre-trained weights and construct a reasoning dataset using their data construction pipeline to maintain consistency with their methodology. To ensure fair evaluation, we use the final checkpoints of all models, including ours, avoiding any form of cherry-picking. This approach allows for a robust comparison and highlights the performance differences across the various models.


\begin{table*}[h]
    \centering
    \vspace{0.3cm}
    \caption{\textbf{Ablation study on visual affordance and textual affordance.} Our experiments demonstrate that both affordance are important for VLA.}
    \label{tab:affordance_ablation}
    \resizebox{0.9\textwidth}{!}{%
    \begin{tabular}{l|c|c|c|c|c}
        \toprule
         & LIBERO-Spatial & LIBERO-Object & LIBERO-Goal & LIBERO-Long & Average \\
        Method / Task & Success Rate ($\uparrow$) & Success Rate ($\uparrow$) & Success Rate ($\uparrow$) & Success Rate ($\uparrow$) & Success Rate ($\uparrow$) \\
        \midrule
        \textbf{CoA-VLA} & \textbf{$85.3 \pm 0.9\%$} & \textbf{$93.1 \pm 1.0\%$} & \textbf{$85.8 \pm 0.9\%$} & \textbf{$55.0 \pm 1.2\%$} & \textbf{$79.8 \pm 0.5\%$}\\
        w/o visual affordance & $84.3 \pm 0.5\%$ &  $91.5 \pm 0.7\%$ & $83.9 \pm 1.0\%$ & $54.6 \pm 1.2\%$ & $78.6 \pm 0.9\%$\\
        w/o textual affordance & $81.6 \pm 0.7\%$ & $89.8 \pm 0.9\%$ & $80.1 \pm 1.0\%$ &  $52.5 \pm 0.9\%$ & $76.0 \pm 0.9\%$\\
        \bottomrule
    \end{tabular}%
    }
    
\end{table*}

\begin{table*}[h]
    \centering
    \vspace{0.3cm}
    \caption{\textbf{Ablation study on dynamic affordance selection.} Removing dynamic affordance selection causes introduction of redundant affordance into the learning process, which cause the model to perform even worse than the baseline without it.}
    \label{tab:dynamic_ablation}
    \resizebox{0.9\textwidth}{!}{%
    \begin{tabular}{l|ccccc|c}
        \toprule
         & LIBERO-Spatial & LIBERO-Object & LIBERO-Goal & LIBERO-Long & Average & Inference  \\
        Method / Task & Success Rate ($\uparrow$) & Success Rate ($\uparrow$) & Success Rate ($\uparrow$) & Success Rate ($\uparrow$) & Success Rate ($\uparrow$) & Speed \\
        \midrule
        \textbf{CoA-VLA} & \textbf{$85.3 \pm 0.9\%$} & \textbf{$93.1 \pm 1.0\%$} & \textbf{$85.8 \pm 0.9\%$} & \textbf{$55.0 \pm 1.2\%$} & \textbf{$79.8 \pm 0.5\%$} & 6Hz\\
        - dynamic affordance selection & $85.1 \pm 0.9\%$ &  $92.4 \pm 1.0\%$ & $85.2 \pm 1.0\%$ & $55.2 \pm 1.1\%$ & $79.5 \pm 1.0\%$ & 1Hz\\
        \bottomrule
    \end{tabular}%
    }
    
\end{table*}

\subsection{Details for LIBERO Simulation}
LIBERO is a robot learning benchmark comprising over 130 language-conditioned manipulation tasks. We follow the setting as in OpenVLA~\cite{openvla} open-sourced code and test on four task suites: LIBERO-Spatial, LIBERO-Goal, LIBERO-Object, and LIBERO-Long.

Each suite includes 10 distinct tasks with 50 demonstrations per task. Each task suite emphasizes unique challenges in imitation learning: LIBERO-Goal features tasks with similar object categories but different goals. LIBERO-Spatial requires policies to adapt to varying spatial arrangements of the same objects. LIBERO-Object keeps the layout consistent while changing the objects. During experimentation, our method uses a static camera, and a wrist-mounted camera all methods are evaluated across 1500 trials in total. We filter out the failure data and increase the image resolution to $224 \times 224$. The affordance data is generated using our proposed pipeline for data in LIBERO. In Table~\ref{tab:libero_comparison}, we directly cite the results of Diffusion Policy, Octo, and OpenVLA from OpenVLA's paper. Therefore, to ensure all methods are evaluated fairly, we evaluated our methods across 500 trials for each task suite, and the reported performance is the average success rate over three random seeds. We use the same test data as in OpenVLA. For the baseline ScaleDP, except for using all camera views, all other implementations kept the same.

\section{More Experiments}
\subsection{Ablation Study on Visual-Textual Affordance}
Our primary contribution lies in the introduction of textual affordances and visual affordances, paired with a novel visual-textual co-injection module designed to synergistically integrate these modalities into policy learning. To validate their individual and combined efficacy, we conduct a systematic ablation study (Table~\ref{tab:affordance_ablation}) on the LIBERO robotic task benchmark. Our key finding is that both textual and visual affordances are critical to model performance. Removing either modality leads to significant degradation in task success rates. While both modalities contribute uniquely, textual affordances exhibit stronger influence on policy optimization. We hypothesize that this stems from language’s inherent capacity to encode task-specific semantics (e.g., "pour-able" or "graspable"), which provides clearer optimization signals compared to visual features that require implicit spatial grounding. These results underscore the importance of our co-injection module, which dynamically balances and fuses multimodal affordances to maximize policy robustness in diverse environments.

\begin{figure}[t]
    \centering
    \includegraphics[width=0.42\textwidth]{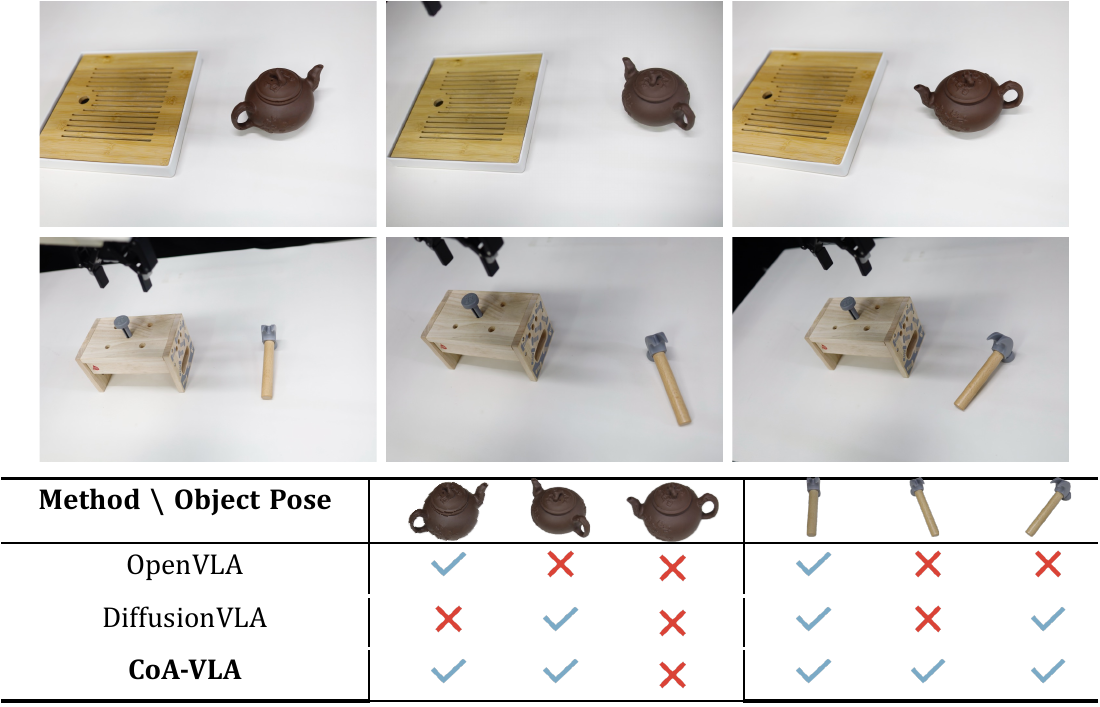}
    \caption{\textbf{Generalization on object pose.}  CoA can pick up objects with unseen poses, benefiting from grasp affordance.}\label{fig:grasp_gen}
\end{figure}

\subsection{Ablation Study on Dynamic Affordance Selection}
Utilizing all affordances can be computationally expensive and time-consuming. Therefore, we introduce a dynamic affordance selection mechanism. This approach focuses on selectively utilizing only the most relevant affordances at each time step. As demonstrated in Table \ref{tab:dynamic_ablation}, our method outperforms a baseline model that employs all affordances indiscriminately.  Surprisingly, using all affordances results in a lower average success rate compared to our dynamic selection approach. We hypothesize that the irrelevant affordances introduce noise during the optimization process, hindering the model's learning ability. To further analyze the impact of dynamic selection, we measured inference speed on an Nvidia 3090 GPU.  We averaged the running time over all tasks, with each task measured across 5 trials.  Our results show that utilizing all affordances significantly impacts inference speed, causing the model to run 6 times slower than our proposed method. This highlights the substantial efficiency gains achieved through dynamic affordance selection

\subsection{Generalization to Unseen Object Pose} 
We assessed CoA-VLA’s ability to generalize to previously unseen object orientations, as illustrated in Figure~\ref{fig:grasp_gen}. Our evaluation focused on two objects: a hammer and a teapot. In the training phase, both objects were consistently presented with their handles oriented vertically relative to the robot. To test the model's generalization capabilities, we introduced novel poses that were absent from the training data, challenging CoA-VLA to grasp these objects in unfamiliar orientations. We observed that CoA-VLA successfully managed most scenarios, demonstrating a remarkable ability to adapt to new object poses even without explicit training on these orientations. In contrast, OpenVLA succeeded only in the simplest cases, struggling with more complex orientations. However, when the objects were positioned horizontally relative to the robot, all models, including CoA-VLA, were unsuccessful in achieving a stable grasp. Despite this limitation, our grasp affordance approach shows promising results, enabling CoA-VLA to handle a wide range of novel object poses.


%% file: sec/table/task_summary.tex
\begin{table}[ht]
\centering

\caption{Summarization for the number of demonstrations and average trajectory length for our real-world tasks.}
\label{tab:task-sum}
\resizebox{0.5\textwidth}{!}{\begin{tabular}{c|c|c|c}
\toprule
\#  & Task               & \# of Demonstrations & Average Trajectory Length \\
\midrule
\hline
1   & PlaceCar                 & 89       & 301.8     \\
2   & PlaceBread                   & 102      & 113.2     \\
3   & NailHammer                  & 80      & 182.8     \\
4   & PourTea                & 91     & 429.4     \\
5   & CleanTrash               & 185     & 114.3     \\
6   & WipeWater             & 62     & 199.9     \\
7 &HangCup  &83        &131.4  \\
\bottomrule
\end{tabular}}
\end{table}